\documentclass[10pt,twocolumn,letterpaper]{article}

\usepackage[pagenumbers]{cvpr} % To force page numbers, e.g. for an arXiv version

\usepackage{graphicx}
\usepackage{amsmath}
\usepackage{amssymb}
\usepackage{booktabs}
\usepackage{subcaption}
\usepackage[normalem]{ulem}
\usepackage[table,xcdraw,dvipsnames]{xcolor}
\usepackage{colortbl}
\usepackage{comment}
\usepackage{times}
\usepackage{multirow}
\usepackage{tabularx}
\usepackage{siunitx}
\usepackage{capt-of}
\usepackage{wrapfig}
\usepackage{lipsum, babel}
\usepackage{pifont}% http://ctan.org/pkg/pifont
\usepackage{soul}
\usepackage{accsupp}
\usepackage[accsupp]{axessibility}

\usepackage[pagebackref,breaklinks,colorlinks]{hyperref}

\usepackage[capitalize]{cleveref}
\crefname{section}{Sec.}{Secs.}
\Crefname{section}{Section}{Sections}
\Crefname{table}{Table}{Tables}
\crefname{table}{Tab.}{Tabs.}
\crefname{equation}{Eq.}{Eqs.}
\Crefname{equation}{Equation}{Equations}

\arrayrulecolor{gray}
\newcommand{\mybullet}{\vspace{0.05cm}\noindent $\bullet$\ }

\def\editing{0}

\if\editing1
    \newcommand{\sachini}[1]{\textcolor{teal}{{#1}}}
    \newcommand{\yasu}[1]{\textcolor{orange}{{#1}}}
    \newcommand{\steven}[1]{\textcolor{purple}{{#1}}}
    \newcommand{\david}[1]{\textcolor{green}{{#1}}}
    \newcommand{\chen}[1]{\textcolor{cyan}{{#1}}}
    \newcommand{\planning}[1]{\textcolor{OliveGreen}{{#1}}}
    \newcommand{\discuss}[1]{\textcolor{Salmon}{{#1}}}
\else
    \newcommand{\sachini}[1]{}
    \newcommand{\yasu}[1]{}
    \newcommand{\steven}[1]{}
    \newcommand{\david}[1]{}
    \newcommand{\chen}[1]{}
    \newcommand{\planning}[1]{}
    \newcommand{\discuss}[1]{#1}
\fi

\newcommand{\mysubsubsection}[1]{\vspace{0.1cm} \noindent {\bf #1}:}
\newcommand{\mysubsubsectionA}[1]{\vspace{0.1cm} \noindent {\bf #1}}

\newcommand{\shorttitle}{NILoc}
\newcommand{\shorttitles}{\shorttitle\ }
\definecolor{darkgreen)}{rgb}{0.0, 0.5, 0.0}
\newcommand{\best}[1]{{\color{orange}{#1}}}
\newcommand{\second}[1]{{\color{cyan}{#1}}}
\def\cvprPaperID{1007} % *** Enter the CVPR Paper ID here
\def\confName{CVPR}
\def\confYear{2022}

\begin{document}

\title{Neural Inertial Localization}
\author{
Sachini Herath$^{1*}$\quad David Caruso$^{2}$\quad Chen Liu$^{2}$\quad Yufan Chen$^{2}$\quad Yasutaka Furukawa$^{1}$
\vspace{3pt} \\
\normalsize $^{1}$Simon Fraser University,
BC, Canada \quad \ 
$^{2}$Reality Labs, Meta, Redmond, USA
}

\twocolumn[{%
\renewcommand\twocolumn[1][]{#1}%
\maketitle
\begin{center}
    \vspace{-3mm}
    \centering
    \includegraphics[width=.93\linewidth]{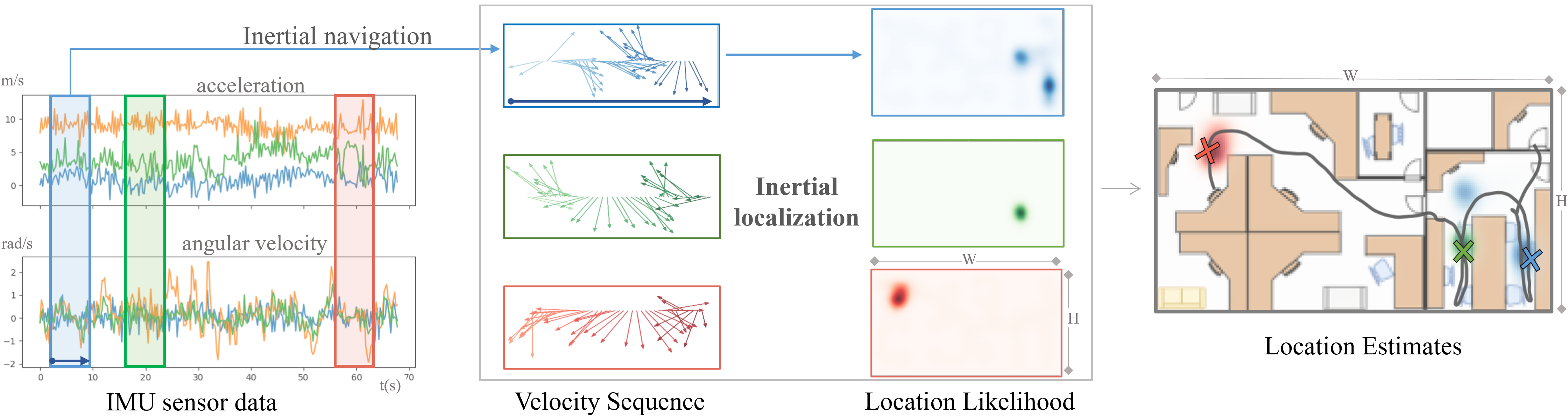}
 \captionof{figure}{From IMU measurements to location estimates. Given inertial sensor data (left), our approach (center) uses a neural inertial navigation technique to find a sequence of velocity vectors; then train a scene-specific transformer-based neural architecture, which maps a velocity sequence to a location likelihood. The figure shows sample localization results from our office dataset. Color shows temporal correspondence between input(left), intermediate representations(middle), and location (right).
 }
 \label{fig:teaser}
\end{center}%
}]

\renewcommand{\thefootnote}{\fnsymbol{footnote}}
\footnotetext[1]{Corresponding author \url{sherath@sfu.ca}. (This work was partially done when Sachini was an intern at Meta.)}
\renewcommand*{\thefootnote}{\arabic{footnote}}

% \maketitle
%%%%%%%%% ABSTRACT
\begin{abstract}
This paper proposes the inertial localization problem, the task of estimating the absolute location from a sequence of inertial sensor measurements. This is an exciting and unexplored area of indoor localization research, where we present a rich dataset with 53 hours of inertial sensor data and the associated ground truth locations.
We developed a solution, dubbed neural inertial localization (\shorttitle) which 1) uses a neural inertial navigation technique to turn inertial sensor history to a sequence of velocity vectors; then 2) employs a transformer-based neural architecture to find the device location from the sequence of velocities.
We only use an IMU sensor, which is energy efficient and privacy preserving compared to WiFi, cameras, and other data sources.
Our approach is significantly faster and achieves competitive results even compared with state-of-the-art methods that require a floorplan and run 20 to 30 times slower.
We share our code, model and data at \url{https://sachini.github.io/niloc}.
\vspace{-4mm}
\end{abstract}

%%%%%%%%% BODY TEXT
\section{Introduction}

Imagine one stands up, walks for 3 meters, turns right, and opens a door in an office. This information might be sufficient to identify the location of the individual.
A recent breakthrough in inertial navigation~\cite{ronin,tlio,idol} allows us to obtain such motion history using an inertial measurement unit (IMU). What is missing is the technology that maps a motion history to a location.
This papers addresses this gap, seeking to open a new paradigm in the localization research, named ``inertial localization'', whose task is to infer the location from a sequence of IMU sensor data. 

Indoor localization is a crucial technology for location-aware services, such as mobile business applications for consumers, %\steven{not clear what this means}, 
entertainment (e.g., Pokemon Go) for casual users, and industry verticals for professional operators (e.g., maintenance at a factory).
State-of-the-art indoor localization systems~\cite{flp}
mostly rely on WiFi, whose infrastructure is ubiquitous thanks to the demands on Internet of Things (IoT). Nevertheless, accuracy of WiFi based localization depends on infrastructure (i.e number of access points) thus cannot scale easily to non-commercial private spaces.

IMU is a powerful 
complementary modality to WiFi, which has proven effective for the navigation task recently~\cite{ronin,tlio,idol}. IMU 1) works anytime anywhere (e.g., inside a pocket/bag/hand); 2) is energy efficient to be an always-on sensor 
and 3) protects the privacy of bystanders.

This paper introduces a novel inertial localization problem as a task of estimating the location from a history of IMU measurements.
The paper provides the first inertial localization benchmark, consisting of 53 hours of motion data and ground-truth locations over 3 buildings.
The paper also proposes an effective solution to the problem, dubbed neural inertial localization (NILoc). NILoc first uses a neural inertial navigation technique~\cite{ronin} to turn IMU sensor data into a sequence of velocity vectors, where the remaining task is to map a velocity sequence to a location.
The high uncertainty in this remaining task is the challenge of inertial localization.
For instance, a stationary motion can be anywhere, and a short forward motion can be at any corridor. 
To overcome the uncertainty, our approach employs a Transformer-based neural architecture~\cite{vaswani2017attention} (capable of encoding complex long sequential data) with a Temporal Convolutional Network (further expanding the temporal capacity by compressing the input sequence length) and an auto-regressive decoder (handling arbitrarily long sequential data).

The contributions of the paper are 3-fold: 1) a novel inertial localization problem, 2) a new inertial localization benchmark, and 3) an effective neural inertial localization algorithm.
We will share our code, models and data.

\section{Related Work}

\begin{table*}[tb]
\centering
\begin{tabular}{|l||lr||crr||cr|}
\hline
\multicolumn{1}{|c||}{\multirow{3}{*}{Building}} & \multicolumn{2}{c||}{Environment} & \multicolumn{3}{c||}{Full Dataset} & \multicolumn{2}{c|}{Test set} \\ \cline{2-8} 
\multicolumn{1}{|c||}{} & \multicolumn{1}{c|}{Dimensions} & \multicolumn{1}{c||}{resolution} & \multicolumn{1}{c|}{\#T (\#S)} & \multicolumn{1}{c|}{duration} & \multicolumn{1}{c||}{length} & \multicolumn{1}{c|}{By sequence} & \multicolumn{1}{c|}{By length} \\
\multicolumn{1}{|c||}{} & \multicolumn{1}{c|}{[$m^2$]} & \multicolumn{1}{c||}{[pixels/m]} & \multicolumn{1}{c|}{} & \multicolumn{1}{c|}{[h]} & \multicolumn{1}{c||}{[km]} & \multicolumn{1}{c|}{\#T (avg. [min])} & \multicolumn{1}{c|}{\#T (100m)} \\ \hline
University A & \multicolumn{1}{l|}{62.8$\times$ 84.4} & 2.5 & \multicolumn{1}{r|}{151 (52)} & \multicolumn{1}{r|}{25.57} & 65.35 & \multicolumn{1}{r|}{25 (12.07)} & 75 \\
University B & \multicolumn{1}{l|}{57.6$\times$147.2} & 2.5 & \multicolumn{1}{r|}{91 ( \  3)} & \multicolumn{1}{r|}{14.64} & 56.93 & \multicolumn{1}{r|}{20 (12.28)} & 60 \\
Office C & \multicolumn{1}{l|}{38.4$\times$ 11.2} & 10.0 & \multicolumn{1}{r|}{81 ( \  1)} & \multicolumn{1}{r|}{12.91} & 21.36 & \multicolumn{1}{r|}{12 (15.48)} & 36 \\ \hline
\end{tabular}
\caption{Inertial localization dataset consists of two university buildings and one office space. The table shows the number of trajectories (\#T), the number of subjects in data collection (\#S) and the length statistics for full dataset and test sets.}
\label{tbl: dataset}
\vspace{-2mm}
\end{table*}

\subsection{Indoor localization}
Outdoor navigation predominately uses satellite GPS. Indoor localization often relies on multiple data sources such as images, WiFi, magnet, or IMU. We review indoor localization techniques based on the input modalities.

\mysubsubsectionA{Image based localization} 
estimates a camera's 6DoF pose from a query image. A classical approach is to detect feature pixels, establish 2D-to-3D or 2D-to-2D correspondences, and solve a Perspective-n-Point (PnP) problem~\cite{rickszeliski_textbook}. The surge of deep learning allows us to learn these steps by an end-to-end network~\cite{yi2016lift}.
Another family of neural architectures directly regress pose parameters~\cite{kendall2015posenet}.
InLoc \cite{taira2018inloc}, a system for image-based indoor localization, reports localization error below 1 meter for 69.9\% of  queries.
While being precise, the image modality suffers from a few major drawbacks for serving mobile applications: a camera needs a direct line of sight, consumes significant amount of battery, and reveals information about bystanders.

\mysubsubsectionA{Wireless localization} based on WiFi or Bluetooth is the current main-stream modality for indoor localization~\cite{sikeridis2018unsupervised, chen2016ibeacon, yang2015wifi}. Wireless receivers work anytime anywhere and WiFi infrastructure is ubiquitous thanks to the ever-growing IoT market demands. The wireless modality is not as precise as the image-based approach and reports a minimum 10 meter error radius\cite{fusion_dhl}.
This paper studies inertial localization as an effective complementary modality. 

\mysubsubsectionA{Activity and magnetic-fields} are other modalities for indoor localization. Activity recognition based on IMU sensor data provides cues on the locations through
a predetermined mapping between activity types and locations~\cite{kamiya2019, zhou2015activity}.
Location specific magnetic-field distortions can also be used to build a localization system through a site survey~\cite{akai2015gaussian, torres2015ujiindoorloc, niuqun2020}.

\mysubsubsectionA{IMU and floorplan} fusion allows classical filtering methods (e.g., particle filter) to perform localization~\cite{thrun2002probabilistic} by using inertial navigation to propagate particles and the floorplan to re-weight particles. 
This approach is sensitive to cumulative sensor errors in inertial navigation.
Correlation between a short motion history (five seconds) and floorplan can provide additional prior to weigh particles~\cite{melamed2021learnedprior}
but requires start location and orientation to initialize the system.
We employ a novel Transformer-based neural architecture that regresses the location from long 
motion history even under severe bending. Our approach does not require a floorplan, which often misses transient objects (e.g., chairs/desks) and needs occasional updates\discuss{, thus provide a compelling alternative}.

\subsection{Inertial navigation} 
Inertial navigation estimates relative motions from IMU sensor data, that is, linear accelerations from an accelerometer and angular velocities from a gyroscope. Deep learning has made significant progress
in recent years, producing 
accurate 3DoF~\cite{ronin} and 6DoF \cite{tlio, idol} trajectories by learning repetitive human body motions with Convolutional Neural Network (CNN)~\cite{resnet} or LSTM\cite{LSTM}.
The main source of error is the bias 
in consumer-grade gyroscopes, accumulating into significant orientation errors, which becomes as large as 20 degrees with 5 minutes of motions even after the calibration. 
Our inertial localization algorithm first uses inertial navigation to estimate a sequence of velocity vectors from IMU data.

\section{Inertial Localization Problem}\label{sec:problem}

Inertial localization is the task of estimating the location of a subject in an environment, solely from a history of IMU sensor data.
There is a training phase and a testing phase \discuss{i.e. without a use of a floorplan or external location information}.
During testing, an input is a sequence of acceleration (accelerometer), angular velocity (gyroscope), and optionally magnetic field (compass) measurements, each of which has 3 DoFs.
An output is position estimations for a given set of timestamps, when ground-truth positions are available.
During training, we have a set of input IMU sensor data and output positions.

\mysubsubsection{Metrics}
The localization accuracy is measured by 
1) the ratio (\%) of correct position estimations within a distance threshold (1, 2, 4, or 6 meters)~\cite{taira2018inloc}; and 2) the ratio (\%) of correct velocity directions within an angular threshold (20 or 40 degrees). The position ratio is the main metric, while the direction ratio measures the temporal consistency. 

\mysubsubsection{Re-localization task extension}
We propose an inertial re-localization task, which is different from inertial localization in that the position $\mathbb{R}^2$ (and optionally the motion direction $SE(2)$) is known apriori. The task represents a scenario where one uses WiFi to obtain a global position once in a few minutes, while re-localizing oneself in-between with an IMU sensor for energy efficiency.

\section{Inertial Localization Dataset}
\label{sec:data}

\begin{figure}[tb]
\centering
\includegraphics[width=.75\linewidth]{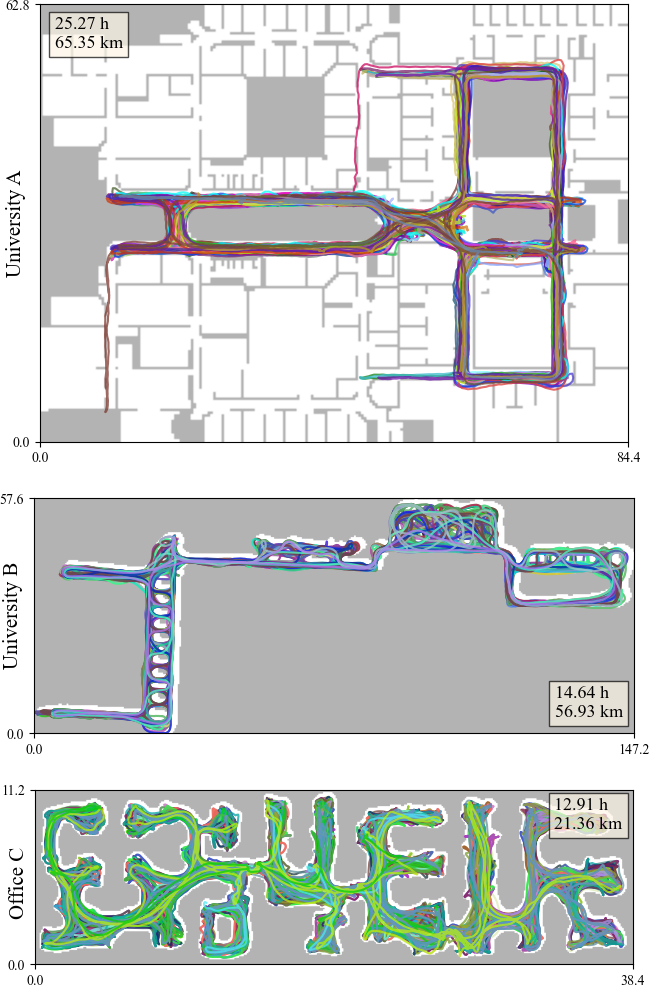}
\caption{Inertial localization dataset contains IMU measurements and ground-truth locations based on Visual Inertial SLAM in three buildings. The ground-truth trajectories are shown here with random colors, overlaid on the respective floorplans. (Image dimensions are in meters).
}
\label{fig:dataset}
\centering
\vspace{-4mm}
\end{figure}

We present the first inertial localization dataset, containing $53$ hours of motion/trajectory data from two university buildings and one office space. Table~\ref{tbl: dataset} summarizes the dataset statistics, while \cref{fig:dataset} visualizes all the ground-truth trajectories overlaid on a floorplan.
Each scene spans a flat floor and the position is given as a 2D coordinate without the vertical displacement. 
If available, a floorplan image is provided for a scene 
for qualitative visualization,
which depicts architectural structures (e.g., walls, doors, and windows) but does not contain transient objects such as chairs, tables, and couches. 

\mysubsubsection{Data collection}
We collect IMU sensor data and ground-truth locations with smartphones.
In the future, AR devices (e.g., Aria glasses by Meta, Spectacles by Snap) will allow collection of ego-centric datatsets with tightly coupled IMU and camera data.
We used two devices in this work; 1) a handheld 3D tracking phone (Google Tango, AsusZenfone AR) with built-in Visual Inertial SLAM capability, producing ground-truth relative motions, where the Z axis is aligned with the gravity; and 2) a standard smartphone, recording IMU sensor data under natural phone handling (e.g. in a pocket, hand or used for calling etc.).  
We utilize Tango Area Description Files \cite{tango} to align ground-truth trajectories to a common coordinate frame then manually align with the floorplan. University A contains data from RoNIN dataset \cite{ronin} aligned manually to a floorplan.
Both IMU sensor data and ground-truth positions are recorded at 200Hz. 

\mysubsubsection{Test sequences}
We randomly select one sixth of the trajectories as the testing data, whose average duration is 13.3 minutes.
We also form a set of short fixed-length sub-sequences for testing
by randomly cropping three sub-sequences (100 meters) from each testing sequence.
 
\shorttitles dataset is de-identified to mask the identity of subjects and does not contain any image or video data.

\section{NILoc: Neural Inertial Localization}
\label{sec:method}

\begin{figure*}[t]
\centering
\includegraphics[width=.95\linewidth]{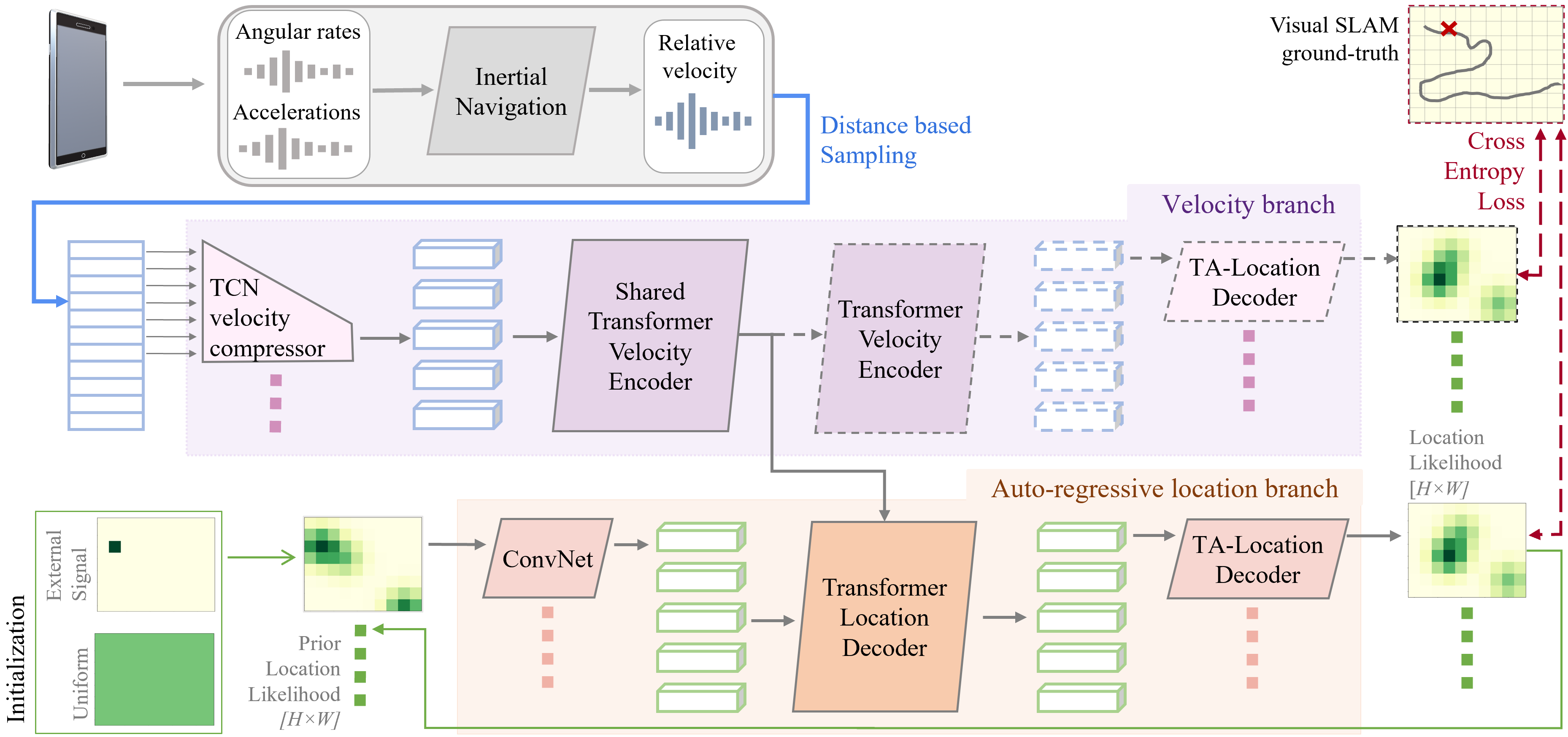}
\caption{Neural inertial localization system diagram.
We use two branch transformer architecture to estimate location likelihood from velocity input. The paths only used in training are shown as dashed lines.
}
\label{fig:network}
\centering
\vspace{-3mm}
\end{figure*}

Instead of regressing locations from IMU measurements, our system \shorttitle\ capitalizes on neural inertial navigation technology~\cite{ronin} that turns a sequence of IMU sensor data to a sequence of velocity vectors, where our core task will be to turn velocity vectors into location estimations.~\footnote{RoNIN ResNet model estimates velocities at the frequency of IMU data. To handle periods of no to little motions (e.g., sitting down), we resample velocities based on the distance of travel. Concretely, we add-up velocity vectors until its length is more than the distance equivalent to one pixel in the location map, then sample one aggregated vector.}

High uncertainty is the challenge in the task. NILoc employs a neural architecture with two Transformer-based network  branches~\cite{vaswani2017attention}, capable of using long history of complex motion data to reduce uncertainty. 
The ``velocity branch'' encodes a sequence of velocity vectors, where a Temporal Convolutional Network compresses temporal dimension to further augment the temporal receptive field. The ``auto-regressive location branch'' encodes a sequence of location likelihoods, capable of auto-regressively producing location estimates on a long horizon.
The network is trained per scene given training data.

The section explains the two branches (\cref{sec:velocity branch,sec:location branch}), the training scheme (\cref{sec:training}), and the data augmentation process (\cref{section:augmentation}), which proves effective in the absence of sufficient training data.

\subsection{Velocity branch}\label{sec:velocity branch}
The branch estimates a location sequence using a history of velocity data. It consists of three network modules:
TCN-based velocity compressor, Transformer velocity encoder, and Translation-aware location decoder.

\mysubsubsection{TCN-based velocity compressor}
Transformer is powerful but memory intensive. We use a temporal convolutional network (TCN)~\cite{TCN} to compress a velocity sequence length by a factor of 10,
allowing us to process longer motion history. In particular, we use a 2-layer TCN with a receptive field of 10 to compress a sequence of 2D velocity vectors $\{v_t\}$ of length T into a sequence of $d$-dimensional{\footnote{Dimension $d$ is set to 288, 470, and 448 for buildings A, B, and C, respectively to be proportional to their floor-areas and resolution.}}
feature vectors  $\{v^\prime_t\}$ of length T/10:
\vspace{-2mm}
\begin{equation*}
    \{v_1, v_2, \cdots \cdots v_T\} \longrightarrow \{v^\prime_1, v^\prime_2, \cdots v^\prime_{T/10}\}.
\vspace{-1mm}
\end{equation*}

\mysubsubsection{Transformer encoder}
Transformer architecture~\cite{vaswani2017attention} takes compressed velocity vectors $\{v^\prime_t\}$ as tokens and initializes each feature vector $f_t$ by concatenating the $d/2$ dimensional trigonometric position encoding of the frame index: 
\vspace{-1mm}
\begin{eqnarray}
f_t &=& \left[ v^\prime_t,\  \{\cos{(w_i t)}\}, \{\sin{(w_i t)}\} \right]\notag \\
w_i &=& 
\exp{\frac{-log(10000) * i}{d'}} \ (i = 1, 2, \cdots d/4)
\notag
\vspace{-1mm}
\end{eqnarray}
$f_t$ is of dimension $d' (=\frac{3}{2} d)$. An
output embedding $e_t$ per token is also a $d'$ dimensional vector encoding the location likelihood.
The encoder has two blocks of self-attention networks. Each block
has 2 standard transformer encoder layers with 8-way multi-head attention.
Feature vectors after the first block are also passed to the other branch (i.e., auto-regressive location branch).

\mysubsubsection{Translation-aware location decoder}
The last module operates on each individual embedding $e_t$. First, $e_t$ is rearranged into an image feature volume (3D tensor~\footnote{The dimensions  (width,height,channels) 
are 24x18x1, 16x44x1, and 14x48x1 for the three scenes A, B, and C, respectively}) 
and up-sampled via a 3-layer fully convolutional decoder with transpose convolutions. The last layer is a ``translation-aware'' $1\times 1$ convolution, whose parameters are not shared across pixels. 
To account for uncertainty, the output location is represented 
as a 2D likelihood map $L_t$ of size $W{\times}H$: $L(x,y)$.~\footnote{The map-extent is determined for each scene by the axis-aligned bounding box of ground-truth locations. We choose the resolution (pixels per meter) so that total number of pixels is around 3 million
(See Table~\ref{tbl: dataset}).}
This translation-aware layer allows the network to easily learn translation-dependent information such as ``people never come to this location'' or ``one always pass through this doorway''.

\subsection{Auto-regressive location branch}\label{sec:location branch}
The location branch combines the velocity features from the velocity branch and prior location likelihoods, which comes from its past inference or an external position information such as WiFi.

The location branch has the same architecture as the velocity branch with two differences. First, instead of a TCN-based velocity compressor, we use a ConvNet to convert each $W{\times}H$ likelihood map into a $d'$-dimensional vector. We use the same trigonometric position encoding (but with dimension $d'$ instead of $d/2$ to match the dimension), which is added to the vector.
Second, we inject velocity features from the velocity branch via cross-attention after every self-attention layer (i.e., before every add-norm layer).
The rest of the architecture is the same.
Note that both branches predict locations, and have different trade-offs (See Sect.~\ref{sec:ablation} for ablation study and discussion).

At inference time, we first evaluate the velocity branch in a sliding window fashion to compute velocity feature vectors. The location branch takes a history of location likelihoods up to 20 frames: $\{L_t, L_{t-1}, \cdots L_{t-19}\}$. $L_0$ encodes external initial location information (e.g., from WiFi) or a uniform distribution if not available. At the output, a node initialized with a likelihood at frame $t'$ will have a likelihood estimate at frame $t'+1$. Therefore, we infer a likelihood up to 20 times for one frame, where we compute the weighted average as the final likelihood by decreasing the weights from 1.0 down to 0.05 from the first inference result to the last.

\subsection{Training scheme}\label{sec:training}

We use a cross entropy loss at both branches. The ground-truth likelihood is a zero-intensity image, except for one pixel at the ground-truth location whose value is 1.0. 
We employ parallel scheduled sampling~\cite{mihaylova2019scheduled} to train the auto-regressive location branch without unrolling recurrent inferences. The process has two steps.
First, we pass GT likelihoods to all the input tokens and make predictions.
Second, we keep the GT likelihoods in the input tokens with probability $r_{teacher}$ (known as a teacher-forcing ratio), while replacing the remaining nodes with the predicted likelihoods.
The back-propagation is conducted only in the second step. $r_{teacher}$ is set to 1.0 in the first 50 epochs, and reduced by $0.01$ after every 5 epochs.

\subsection{Synthetic data generation}\label{section:augmentation}
The Transformer architecture requires a large amount of training data.~\footnote{The COVID pandemic further makes the data collection challenging.}
We crop data over different time windows to augment training samples. However, in the absence of sufficient training data, we use the following three steps to generate more training samples synthetically: 1) Compute a likelihood map of training trajectories (i.e., where they pass through); 2) Randomly pick a pair of locations from high likelihood areas; and 3) Solve an optimization problem to produce a trajectory that is smooth and pass through the area of high likelihood. Given a synthesized trajectory, we sample velocity vectors based on the distance of travel as in our preprocessing step, which are directly passed to the TCN-based velocity compressor during training.
All the steps are standard heuristics and we refer the details to the supplementary.

\section{Experimental Results}

\begin{table*}[tbh]
\begin{subtable}{\linewidth}
\centering
\begin{tabular}{|l|l||rrrrrr||rrrrrr||c|}
\hline
\multirow{3}{*}{\begin{tabular}[c]{@{}l@{}}Buil-\\ ding\end{tabular}} & \multirow{3}{*}{Meth.} & \multicolumn{6}{c||}{Fixed short sequence (100 m)} & \multicolumn{6}{c||}{Full test sequence} & \multicolumn{1}{c|}{\multirow{3}{*}{\begin{tabular}[c]{@{}c@{}}run time\\cpu/gpu\\ (sec) $\downarrow$\end{tabular}}} \\ \cline{3-14}
 &  & \multicolumn{4}{c|}{SR(\%) at distance $\uparrow$} & \multicolumn{2}{c||}{SR(\%) at A  $\uparrow$} & \multicolumn{4}{c|}{SR(\%) at distance  $\uparrow$} & \multicolumn{2}{c||}{SR(\%) at A $\uparrow$} & \multicolumn{1}{c|}{} \\ \cline{3-14}
 &  & \multicolumn{1}{c}{1m} & \multicolumn{1}{c}{2m} & \multicolumn{1}{c}{4m} & \multicolumn{1}{c|}{6m} & \multicolumn{1}{c}{$20^{\circ}$} & \multicolumn{1}{c||}{$40^{\circ}$} & \multicolumn{1}{c}{1m} & \multicolumn{1}{c}{2m} & \multicolumn{1}{c}{4m} & \multicolumn{1}{c|}{6m} & \multicolumn{1}{c}{$20^{\circ}$} & \multicolumn{1}{c||}{$40^{\circ}$} & \multicolumn{1}{c|}{} \\ \hline\hline
\multirow{3}{*}{A} & PF & 1.8 & 6.7 & 11.9 & \multicolumn{1}{r|}{15.4} & 21.3 & 32.2 & 6.5 & 16.8 & 22.8 & \multicolumn{1}{r|}{26.2} & 31.5 & 42.5 & \second{0.6}\ /\ 7.7 \\
 & CRF & \second{15.0} & \best{32.5} & \best{46.3} & \multicolumn{1}{r|}{\best{53.6}} & \best{61.7} & \best{70.5} & \second{14.2} & \second{31.9} & \second{47.0} & \multicolumn{1}{r|}{\second{54.7}} & \second{53.0} & \second{61.0} & 9.5\ /\ \second{3.7} \\
 & Ours & \best{16.7} & \second{28.9} & \second{38.8} & \multicolumn{1}{r|}{\second{44.6}} & \second{46.8} & \second{54.1} & \best{23.4} & \best{44.8} & \best{62.6} & \multicolumn{1}{r|}{\best{69.5}} & \best{65.6} & \best{74.8} & \best{0.3\ /\ 0.1} \\ \hline\hline
\multirow{3}{*}{B} & PF & 1.0 & 3.8 & 7.0 & \multicolumn{1}{r|}{9.0} & 17.0 & 27.7 & 6.4 & 16.8 & 28.6 & \multicolumn{1}{r|}{34.0} & 38.1 & 51.6 & \second{1.8\ /\ 1.4} \\
 & CRF & \second{12.4} & \second{33.6} & \second{48.7} & \multicolumn{1}{r|}{\second{53.7}} & \second{62.0} & \second{65.7} & \second{18.4} & \second{49.8} & \second{68.6} & \multicolumn{1}{r|}{\second{71.5}} & \second{71.8} & \second{77.2} & 18.8\ /\ 5.4 \\
 & Ours & \best{47.6} & \best{69.3} & \best{74.5} & \multicolumn{1}{r|}{\best{77.3}} & \best{67.9} & \best{75.1} & \best{49.4} & \best{73.1} & \best{80.1} & \multicolumn{1}{r|}{\best{82.0}} & \best{72.7} & \best{80.7} & \best{1.2\ /\ 0.2} \\ \hline\hline
\multirow{3}{*}{C} & PF & 19.7 & 30.9 & \second{46.0} & \multicolumn{1}{r|}{\second{58.6}} & 21.8 & 38.2 & 18.3 & 28.9 & 43.8 & \multicolumn{1}{r|}{55.2} & 21.0 & 38.0 & \second{4.3\ /\ 4.2} \\
 & CRF & \second{26.3} & \second{36.2} & 43.7 & \multicolumn{1}{r|}{52.1} & \second{31.3} & \second{46.3} & \second{44.3} & \second{60.5} & \second{72.1} & \multicolumn{1}{r|}{\second{80.5}} & \second{44.4} & \second{64.9} & 38.1\ /\ 16.8 \\
 & Ours & \best{69.9} & \best{78.1} & \best{83.4} & \multicolumn{1}{r|}{\best{87.2}} & \best{51.8} & \best{67.4} & \best{72.9} & \best{80.5} & \best{85.2} & \multicolumn{1}{r|}{\best{89.1}} & \best{53.4} & \best{69.7} & \best{2.4\ /\ 0.7} \\ \hline
\end{tabular}
\caption{Inertial Localization}
\label{tbl:results:localization}
\vspace{-1mm}
\end{subtable}\par

\bigskip

\begin{subtable}{\linewidth}
\centering
\begin{tabular}{|l|l|rrrrrr||rrrrrr||r|}
\hline
\multirow{3}{*}{Task} & \multirow{3}{*}{Meth.} & \multicolumn{6}{c||}{Fixed short sequence (100 m)} & \multicolumn{6}{c||}{Full test sequence} & \multicolumn{1}{c|}{\multirow{3}{*}{\begin{tabular}[c]{@{}c@{}}run time\\cpu/gpu\\ (sec) $\downarrow$\end{tabular}}} \\ \cline{3-14}
 &  & \multicolumn{4}{c|}{SR(\%) at distance $\uparrow$} & \multicolumn{2}{c||}{SR(\%) at A  $\uparrow$} & \multicolumn{4}{c|}{SR(\%) at distance  $\uparrow$} & \multicolumn{2}{c||}{SR(\%) at A $\uparrow$} & \multicolumn{1}{c|}{} \\ \cline{3-14}
 &  & \multicolumn{1}{c}{1m} & \multicolumn{1}{c}{2m} & \multicolumn{1}{c}{4m} & \multicolumn{1}{c|}{6m} & \multicolumn{1}{c}{$20^{\circ}$} & \multicolumn{1}{c||}{$40^{\circ}$} & \multicolumn{1}{c}{1m} & \multicolumn{1}{c}{2m} & \multicolumn{1}{c}{4m} & \multicolumn{1}{c|}{6m} & \multicolumn{1}{c}{$20^{\circ}$} & \multicolumn{1}{c||}{$40^{\circ}$} & \multicolumn{1}{c|}{} \\ \hline\hline
\multirow{3}{*}{\begin{tabular}[c]{@{}l@{}}Reloc\\ $\mathbb{R}^2$\end{tabular}} & PF & 21.4 & 40.5 & 60.3 & \multicolumn{1}{r|}{69.9} & 48.9 & 64.1 & 19.7 & 39.6 & 57.0 & \multicolumn{1}{r|}{63.8} & 48.7 & 63.0 & \second{4.7 / 6.7} \\
 & CRF & \second{33.2} & \second{59.7} & \best{78.9} & \multicolumn{1}{r|}{\best{87.7}} & \best{71.7} & \best{83.8} & \second{31.7} & \second{60.3} & \best{79.4} & \multicolumn{1}{r|}{\best{86.7}} & \best{71.7} & \best{84.3} & 21.3 / 8.8 \\
 & Ours & \best{50.9} & \best{69.3} & \second{77.7} & \multicolumn{1}{r|}{\second{82.0}} & \second{65.0} & \second{74.9} & \best{50.8} & \best{69.3} & \second{78.7} & \multicolumn{1}{r|}{\second{82.6}} & \second{65.5} & \second{76.2} & \best{1.3 / 0.3} \\ \hline\hline
\multirow{4}{*}{\begin{tabular}[c]{@{}l@{}}Reloc\\ $SE(2)$\end{tabular}}
% & RoNIN & 9.4 & 23.1 & 46.8 & \multicolumn{1}{r|}{65.6} & \second{68.9} & \best{89.8} & 3.9 & 10.5 & 25.7 & \multicolumn{1}{r|}{40.7} & 52.1 & 75.9 & - \\
& PF & 22.9 & 41.5 & 62.8 & \multicolumn{1}{r|}{73.7} & 51.5 & 68.2 & 15.1 & 30.8 & 47.0 & \multicolumn{1}{r|}{54.9} & 41.8 & 56.0 & \second{2.1} / 6.7 \\
 & LP & 9.7 & 27.1 & 55.3 & \multicolumn{1}{r|}{70.2} & 49.2 & 69.8 & 4.0 & 13.2 & 29.5 & \multicolumn{1}{r|}{40.5} & 36.9 & 54.3 & 7 .0 / \second{2.7} \\
 & CRF & \second{36.5} & \second{64.4} & \best{82.7} & \multicolumn{1}{r|}{\best{90.6}} & \best{74.2} & \best{86.6} & \second{31.9} & \second{61.0} & \best{79.8} & \multicolumn{1}{r|}{\best{86.9}} & \best{72.1} & \best{85.1} & 21.4 / 8.8 \\
 & Ours & \best{52.8} & \best{71.1} & \second{79.4} & \multicolumn{1}{r|}{\second{83.4}} & \second{66.7} & \second{76.6} & \best{51.4} & \best{70.1} & \second{79.6} & \multicolumn{1}{r|}{\second{83.8}} & \second{66.5} & {77.4} & \best{1.3 / 0.3} \\ \hline
\end{tabular}
\caption{Inertial Re-Localization:
% with location  $\mathbb{R}^2$ and 
average metrics across 3 buildings. (see supplementary for results per building)}
\label{tbl:results:relocalization}
\vspace{-1mm}
\end{subtable}
\caption{\shorttitles achieves competitive accuracy at significantly lower run time. We compare \shorttitles (ours) with a 
% \discuss{inertial tracking algoithm: RoNIN \cite{ronin} and} 
three methods that require a floorplan as input: Particle Filter (PF), Learned Prior (LP) and Conditional Random Fields (CRF). We report success rate (SR) at a given error distance threshold and angle (A) threshold. Run time is the average CPU or GPU time per 1 min of motion sequence. The best and second best results per column are shown in \best{orange} and \second{cyan}, respectively.}
\label{tbl:results}
% \vspace{-2mm}
\end{table*}

\begin{figure*}[tbh]
\includegraphics[width=.97\textwidth]{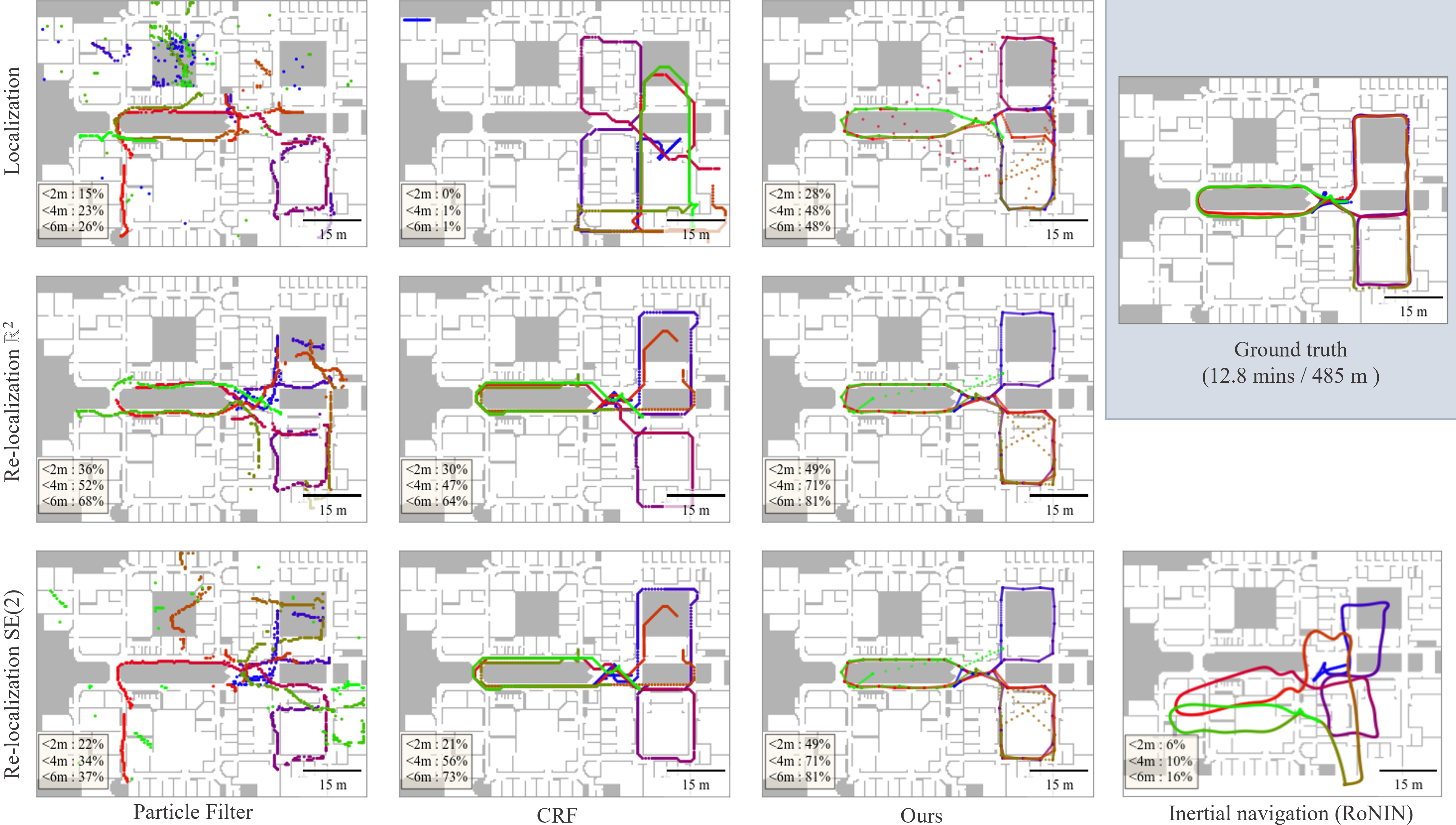}
\caption{Qualitative visualizations: For one trajectory from building A, we show results by the top three methods (columns) for one localization and two re-localization tasks (rows). Particle filter and CRF require a floorplan in addition to IMU input. The color gradient (blue $\rightarrow$ red $\rightarrow$ green) encodes time. We mark the physical dimension of each sequence and report success rate (\%) at distance thresholds 2,4, and 6 meters. See the supplementary for more visualizations.}
\label{fig:viz}
\centering
% \vspace{-1mm}
\end{figure*}

\begin{figure*}[tbh]
\centering
\includegraphics[width=.97\linewidth]{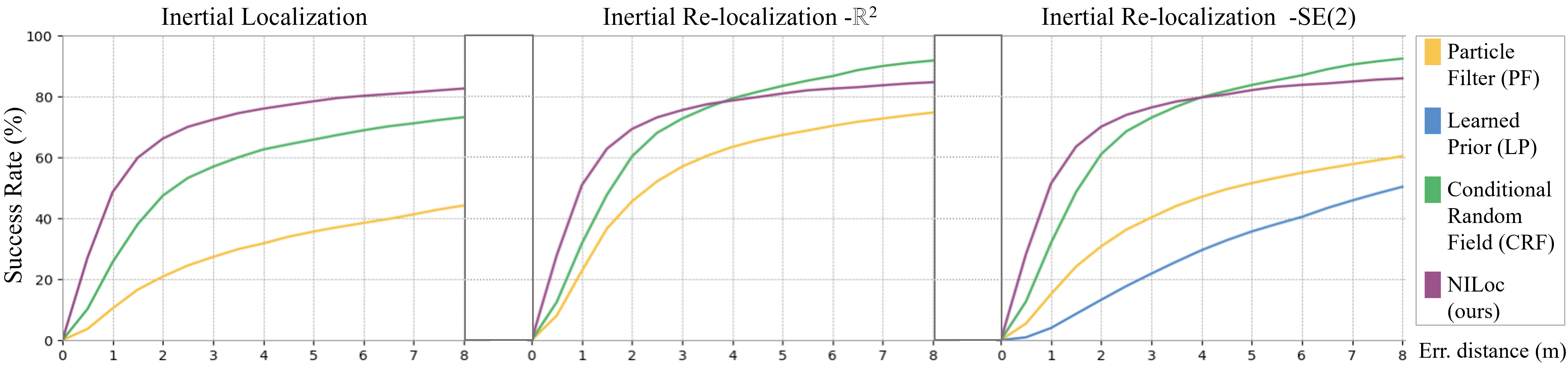}
\caption{The figure plots the success rate metric for the three tasks, while varying the error distance threshold. All baseline except ours require a floorplan as input in addition to IMU. The average score over the three buildings are reported.
}
\label{fig:cdf}
% \vspace{-2mm}
\end{figure*}

\begin{table*}[tbh]
\centering
\begin{tabular}{l|rr|rr|rr|r|}
\cline{2-8}
\multicolumn{1}{c|}{} & \multicolumn{2}{c|}{Localization} & \multicolumn{2}{c|}{Reloc $\mathbb{R}^2$} & \multicolumn{2}{c|}{Reloc $SE(2)$} & \multicolumn{1}{c|}{\multirow{2}{*}{\begin{tabular}[c]{@{}c@{}}Model\\ param.s\end{tabular}}} \\
\multicolumn{1}{r|}{SR(\%) at distance $\rightarrow$} & 2m & 4m & 2m & 4m & 2m & 4m & \multicolumn{1}{c|}{} \\ \hline
\multicolumn{1}{|l|}{w/o \shorttitles (RoNIN \cite{ronin} only)} & - & - & - & - & 10.5 & 25.7 & - \\
% \multicolumn{1}{|l|}{w/o loss on the location branch} & 21.2 & 32.4 & - & - & - & - & 4.6M \\
\multicolumn{1}{|l|}{w/o loss on the velocity branch} & 14.8 & 22.9 & 16.1 & 24.3 & 16.1 & 24.3 & 7.5M \\
\multicolumn{1}{|l|}{w/o velocity compressor} & 3.6 & 7.6 & 6.7 & 12.3 & 7.2 & 12.0 & 10.2M \\
\multicolumn{1}{|l|}{w/o TA-location decoder (FC)} & 5.3 & 10.4 & 6.8 & 13.3 & 8.6 & 14.7 & 211.0M \\
\multicolumn{1}{|l|}{w/o TA-location decoder (CNN)} & 39.5 & 58.3 & 48.8 & 68.6 & 51.1 & 71.1 & 10.2M \\ \hline
\multicolumn{1}{|l|}{Ours (velocity branch output)} & 52.5 & 72.1 & - & - & - & - & 10.5M \\
\multicolumn{1}{|l|}{Ours} & 44.8 & 62.6 & 54.1 & 70.4 & 56.0 & 73.0 & 10.5M \\ \hline
\end{tabular}
\caption{Ablation study. The first row is a inertial navigation algorithm. Next four rows are results after dropping one technical component from our full system.  For the 4rd and the 5th rows, we dropped the translation-aware location decoder and replace by either a fully connected layer (FC) or a fully convolutional decoder (CNN). The last two rows compare predictions by the velocity branch and the location branch, where the latter is the default prediction reported as our result in the other places.
%: improvement from our architecture and data processing pipeline. 
The success rate (\%) at two distance thresholds (m) on building A are the metrics.}
\label{tbl:ablation}
\end{table*}

\subsection{Baseline methods}
To our knowledge, no prior work addresses indoor localization from IMU data alone. Therefore, we compare with the following three techniques that fuse IMU and floorplan.~\footnote{Movements are governed by architectural structures (e.g., walls and rooms) in University A, where a building blueprint is used as a floorplan. For University B and Office C, transient objects such as chairs and desks play a greater role, but do not show up in blueprints. Therefore, we take the likelihood maps from the synthetic data generation process (See Sect.~\ref{section:augmentation}) and binary-threshold them as floorplans.} Note that our method is the only \textit{inertial localization} that uses IMU data alone without the floorplan images.
We briefly explain the three techniques.

\mybullet \textbf{Particle Filter (PF)} maintains a set of particles, each of which stores the location, heading direction, and bias/scale error correction terms. Starting from a Gaussian distribution around the given initial location or a uniform distribution otherwise, the system updates the states of the particles based on the inertial navigation result and the floorplan information (i.e., down-weight particles if outside the walkable region). We take the particle closest to the weighted median x/y coordinate as the location prediction.

\mybullet \textbf{Learned Prior (LP)\cite{melamed2021learnedprior}}
is also a particle filter based approach while using deep networks to help update particle weights. A location likelihood, computed by the dot product between floorplan features extracted by UNet \cite{unet} and motion features extracted by LSTM \cite{LSTM}, is used to weight the particles.
We use our local implementation as the code is not available.
Note that this method requires the initial location and the orientation %(fails badly without the initialization in our experiments), 
and is evaluated only for the $SE(2)$ re-localization task.

\mybullet \textbf{Conditional Random Field (CRF)} is based on a state-of-the-art map-matching system~\cite{xiao2014lightweight}, which computes a reach-ability graph from a floorplan, uses inertial navigation results to transition between graph nodes, and uses Viterbi algorithm to backtrack and determine location. 
We modified the system in a few ways to better adapt to our task: 1) use the RoNIN result as the velocity input; 2) increasing the search neighborhood by a factor of 1.5 to handle scale inaccuracies in inertial navigation; and 3) do periodic back-propagation within a fixed window to be comparable with other near-real time baselines.

\subsection{Implementation details}
We have implemented the proposed system in Pytorch-Lightning~\cite{Falcon_PyTorch_Lightning_2019}. For training, we have used AdamW optimizer~\cite{adamw}. The learning rate has been initialized to $0.0001$ with 30 epoch warm-up \cite{goyal2017accurate}, and reduced by a factor of $0.75$ after every 10 epochs when the validation loss does not decrease. 
We use random one sixth of the training trajectories as a hold-out validation. 
In the main experiments we combine real and synthetic data for training. Specifically, we train on the combined set 
until convergence ($700$ epochs) and fine-tune only with real dataset ($400$ epochs). The total training time is approx. 3 days on GeForce RTX 2080 Ti and 1 day on NVIDIA Tesla V100 GPUs. For RoNIN software, we have downloaded the trained model from the official website~\cite{ronin}.
Note that unless otherwise noted, our results indicate the predictions by the auto-regressive location branch.~\footnote{For the localization task, we initialize with uniform  distribution of particles or location likelihood. For the re-localization task with $\mathbb{R}^2$ (resp. $SE(2)$), we initialize particles from a Gaussian distribution around ground truth location and uniform distribution for heading (resp. Gaussian distribution around location and heading), or provide ground-truth likelihood map of the first frame (resp. the first two input frames).}

For the baseline methods, we use Pytorch\cite{Falcon_PyTorch_Lightning_2019} and CuPy \cite{cupy_learningsys2017} to implement both on CPU and GPU.
For being fair, we use the same floorplan resolution and the same distance-based sampling to extract velocities.
A grid search is used to find hyper parameters for all baselines.

\subsection{Evaluation}

\Cref{tbl:results:localization} is our main result, quantitative evaluations on the localization task for the three buildings separately. \shorttitles achieves the best results in most entries. The only exception is against CRF for building A for fixed short sequences. Note that CRF uses floorplan information while our input is only IMU. CRF is computationally intensive, 30 times slower than our method, involving even a dynamic programming to effectively search for all possible alignments with a floorplan exhaustively. 

\Cref{tbl:results:relocalization} shows the results on the re-localization task, averaged over the three buildings. All methods consistently improves performance with more initialization information. While NILoc achieves the best results for lower distance thresholds (i.e., often extremely accurate), CRF performs better overall at the sacrifice of its intensive computational expenses and the requirement on the floorplan image.

Figure~\ref{fig:cdf} observes the same results, plotting the distance-based success rate over a range of thresholds (averaged over the three buildings).
Particle filter exhibits poor performance, where the major limitation comes from its inability to handle cumulative drifts in the inertial navigation trajectories.
Learned prior combines particle filter and neural networks that learn to associate motions with locations. 
However, they only encode 5 seconds of motion data with LSTM and ConvNet, not long enough to break the ambiguity.
\shorttitles takes roughly a minute of motion data with powerful Transformer based architecture, overcoming the uncertainties. Another observation is that all the baseline methods explicitly integrate velocities to update the location information. \shorttitles does not bake-in this integration formula and relies completely on learning to relate velocities to locations, which could potentially make our approach more robust against cumulative drifts by inertial navigation.

Figure~\ref{fig:viz} provides qualitative visualizations of one trajectory from building A. 
Particle filter and CRF estimate locations at roughly 200Hz, while NILoc is roughly 20Hz. We interpolate NILoc locations in plotting the trajectories, where streaks of points in the figure are interpolation artifacts at discontinuous predictions.
As shown by the inertial navigation trajectory at the bottom right, this is not an easy task where the trajectory suffers from significant cumulative drifts. Nonetheless, our system is capable of inferring correct locations for most of the frames.

\subsection{Ablation study}
\label{sec:ablation}
Table~\ref{tbl:ablation} is an ablation study, assessing the contributions of various technical components in our system. \discuss{The first row compares against a state-of-art inertial navigation method which ours and all baselines outperform (also see \cref{fig:viz}).} The next four rows show the distance-based success rate while dropping four components one by one from our main system. 
The table shows that it is important to train the network with losses on both branches. 
The second row is particularly interesting. Both the location and the (first half of the) velocity branches are trained with the loss only at the location branch, whose performance drops significantly.
The third row indicates the challenge of high uncertainty in the inertial navigation task. The success rates even drop to a single digit, when the input motion history becomes 10 times less without the TCN based compressor.
The last two rows compare the predictions by the velocity branch and the location branch of our full system. The velocity branch does not take previous location likelihoods and cannot solve re-localization tasks. However, for the localization tasks, it outperforms the location branch with a clear margin, while being twice as computationally efficient.

\section{Limitations and Future Work}

There are two major failure modes in our approach. First, in an open space (e.g., an atrium), human motions tend not to follow patterns, where any location could be an answer. Second, in the presence of symmetries or repetitions, multiple locations would be equally likely. \discuss{Our method is designed not to use any future frame information after the first input window so that it can be deployed as a real-time system, where predictions may jump abruptly under high uncertainties.}
Our future work is to exploit body motion signals that are captured in IMU but are currently discarded by inertial navigation and distance-based velocity sampling to overcome the uncertainty. For instance, IMU signals differ when one opens a door, washes hands, or orders a coffee, which provides effective cues in localizing the position. 
Please refer to the supplementary for more qualitative visualizations (i.e., location trajectories as static images and videos for more samples in more buildings) and quantitative ablation studies (e.g., w/o synthetic data, w/o scheduled sampling, or comparison against TCN as a backbone instead of Transformer).
We share our code, models and data to promote further research in the space of inertial localization.

\mysubsubsection{Societal impact} 
Inertial localization could be a critical component for indoor GPS. A mobile app might become capable of recording location history 24/7 anywhere in indoor spaces,
which tend to be more private than outdoors (e.g., inside a house or a rest room). A smartphone developer should understand the impact of giving IMU sensor data and define appropriate access controls for the apps.
On the positive side, the inertial location could happen on-device, which would allow a higher degree of privacy control compared to other data modalities.

% \clearpage
\footnotesize{\mysubsubsection{Acknowledgements} 
The research is supported by NSERC 8 Discovery Grants, NSERC Discovery Grants Accelerator Supplements, and DND/NSERC Discovery Grant Supplement. We thank Weilian Song, Saghar Irandoust and Fuyang Zhang for their contribution to dataset. 

}

% \clearpage

%%%%%%%%% REFERENCES
{\small
\bibliographystyle{ieee_fullname}
\bibliography{egbib}
}

% \clearpage

% \input{supplementary/arxiv_supplementary}

\end{document}

\title{Neural Inertial Localization\\Supplementary Material}

\author{
Sachini Herath$^{1}$\quad David Caruso$^{2}$\quad Chen Liu$^{2}$\quad Yufan Chen$^{2}$\quad Yasutaka Furukawa$^{1}$
\vspace{3pt} \\
\normalsize $^{1}$Simon Fraser University,
BC, Canada \quad \ 
$^{2}$Reality Labs, Meta, Redmond, USA
}

\maketitle

The supplementary document provides: 

\vspace{0.1cm}

\noindent $\bullet$ Algorithmic details 
% on synthetic data generation
(\cref{sec:supple:algo});

\begin{figure*}[p]
\centering
\includegraphics[width=.92\textwidth]{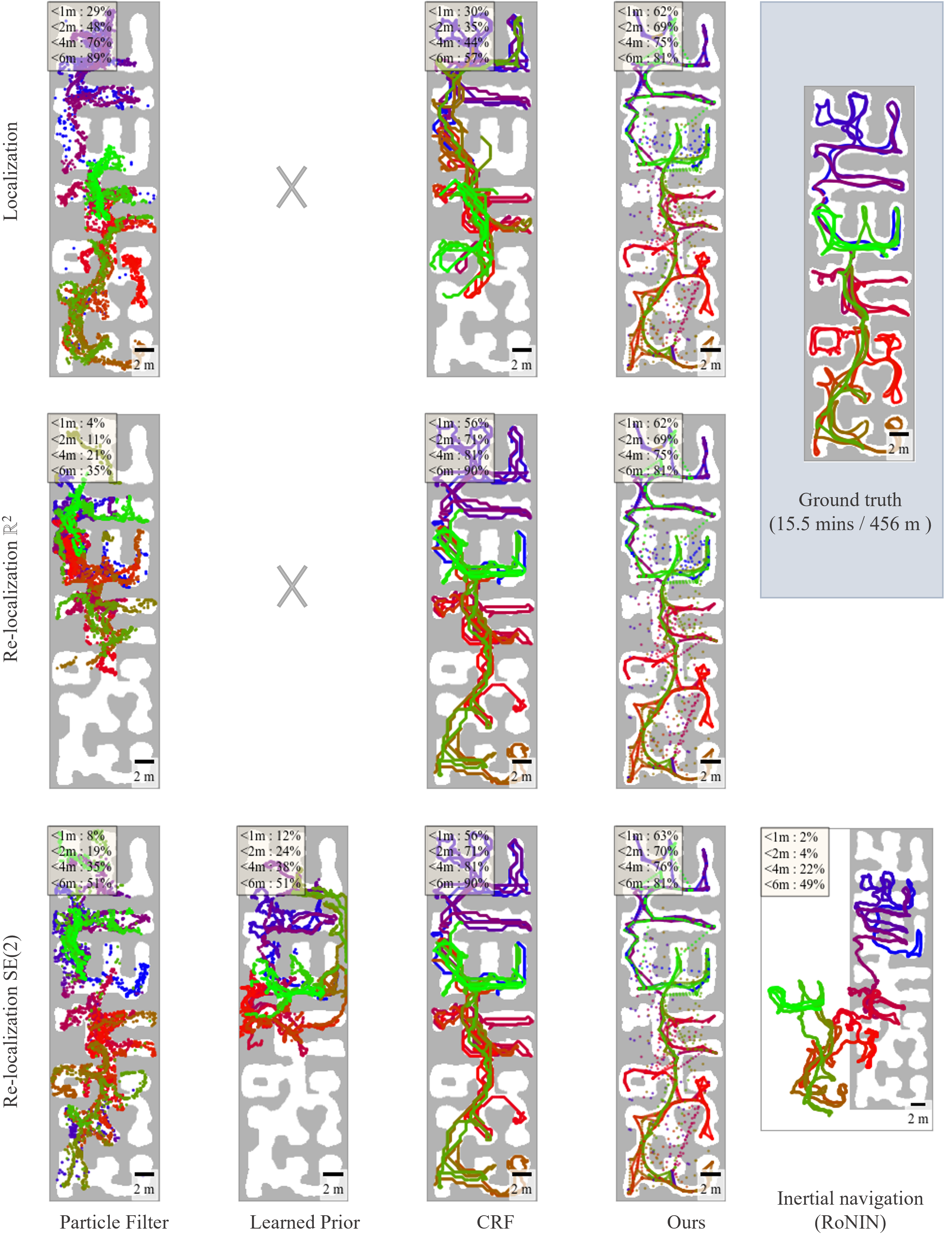}
\caption{Qualitative visualizations: For one trajectory from Office C, we show results by the four methods (columns) for one localization and two re-localization tasks (rows). Particle filter, learned prior and CRF require a floorplan in addition to IMU input. The color gradient (blue $\rightarrow$ red $\rightarrow$ green) encodes time. We mark the physical dimension of each sequence and report success rate (\%) at distance thresholds 1, 2,4, and 6 meters.}
\label{fig:supple:vizB}
\centering
\end{figure*}

\begin{figure*}[p]
\centering
\includegraphics[width=.92\textwidth]{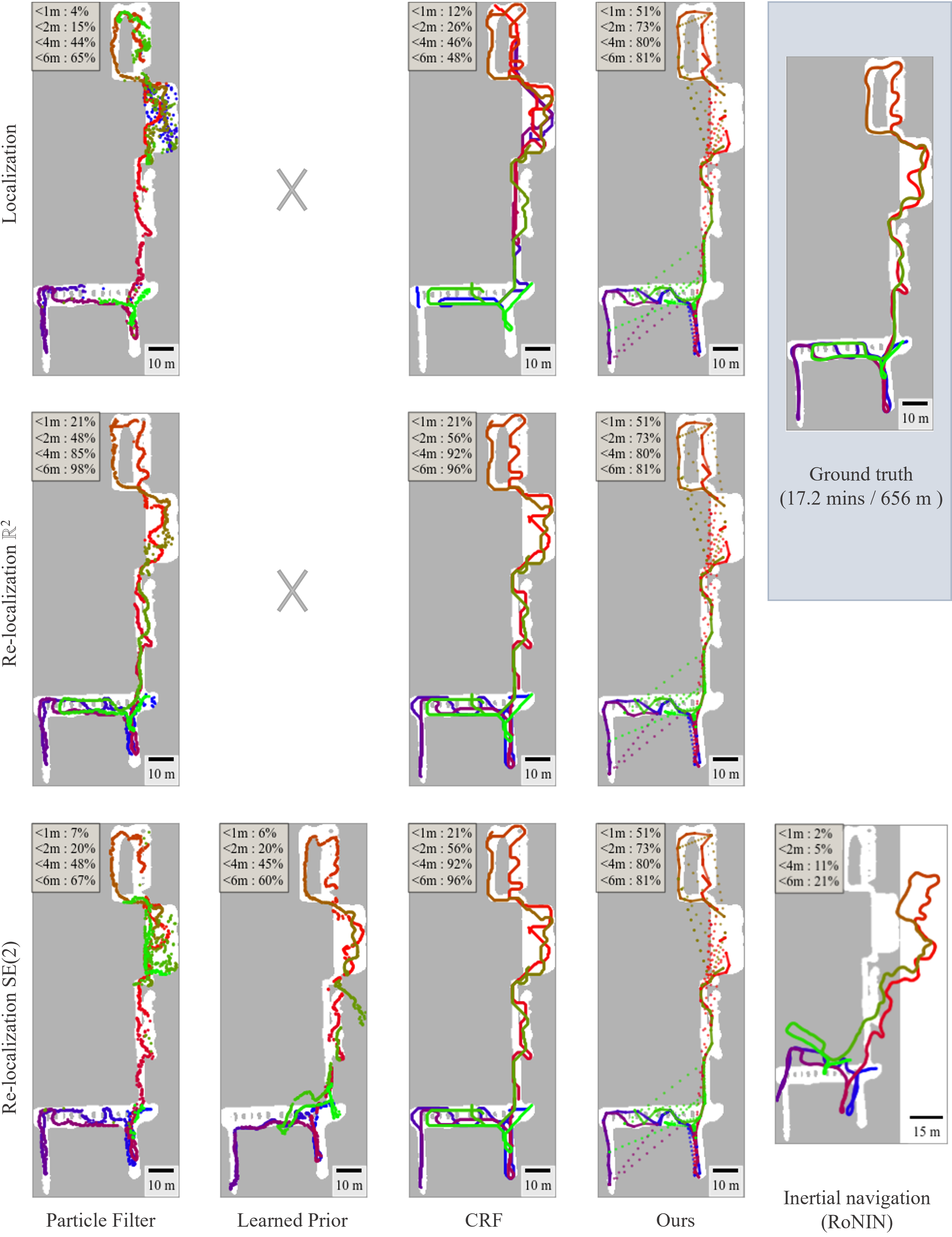}
\caption{Qualitative visualizations: For one trajectory from building B, we show results by the four methods (columns) for one localization and two re-localization tasks (rows). Particle filter, learned prior and CRF require a floorplan in addition to IMU input. The color gradient (blue $\rightarrow$ red $\rightarrow$ green) encodes time. We mark the physical dimension of each sequence and report success rate (\%) at distance thresholds 1, 2, 4, and 6 meters.}
\label{fig:supple:vizC}
\centering
\end{figure*}

\noindent $\bullet$ More qualitative visualization (\cref{fig:supple:vizB} and \ref{fig:supple:vizC});

\begin{figure*}[tb]
\centering
\includegraphics[width=\textwidth]{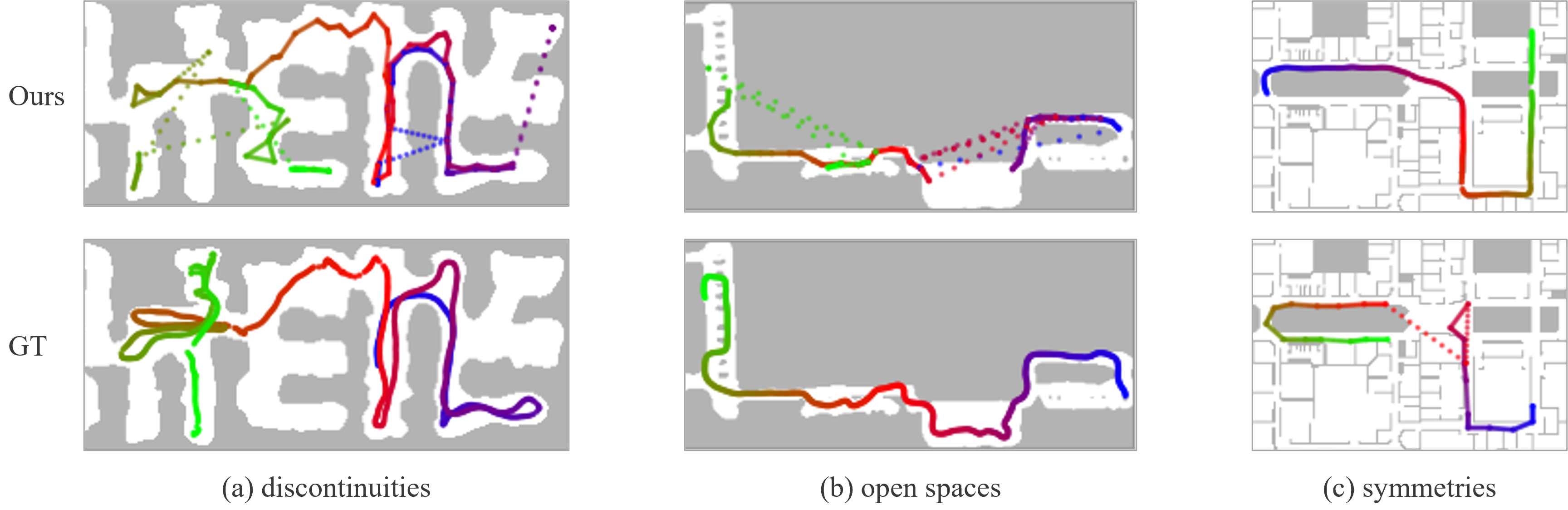}
\caption{Failure cases: Three common failure modes in our approach are shown with our prediction for localization task and ground-truth (GT) trajectory segments. The color gradient (blue $\rightarrow$ red $\rightarrow$ green) encodes time.}
\label{fig:supple:failure}
\centering
\end{figure*}

\noindent $\bullet$ Failure cases (\cref{fig:supple:failure});

\noindent $\bullet$ Architecture specification (\cref{tbl:supple:network});

\noindent $\bullet$ Quantitative evaluations on re-localization task (\cref{tbl:supple:reloc});

\noindent $\bullet$ More ablation studies (\cref{tbl:supple:ablation}).

\vspace{0.1cm}

Please also refer to our supplementary video, which shows predicted likelihoods and motion trajectories for each method in animtaion.

\section{Algorithmic details}\label{sec:supple:algo}
We discuss the details of our synthetic data generation process and data augmentation during training, outlined in section 5.4. 

\subsection{Synthetic data generation}
When a floorplan image is not available, we compute the walklable region or a floorplan ($F_{map}$) by 1) counting the number of times training trajectories pass through at each pixel; 2) clamping the count to be 2 at the maximum; 3) applying Gaussian smoothing ($\sigma=1.0$ pixels); and 4) binary-thresholding the map with a threshold of $0.5$.

We generate a synthetic motion trajectory from the floorplan ($F_{map}$) by picking start and goal positions randomly, and finding a shortest path between them, while we define a local neighborhood system to be a 11x11 square region.

Next, in order to make the trajectories look realistic, we apply B-spline smoothing then solve an optimization problem so that the trajectories pass through near the center of corridors and passages. Note that when synthetic trajectories are used in training, to minimize the gap between synthetic data and real data, we also apply the B-spline smoothing to the real trajectory before passing the velocity vectors to the velocity branch.

\mysubsubsection{B-spline smoothing}\label{sec:bspline_smoothing}
We approximate synthetic trajectory by fitting a b-spline curve~\cite{dierckx1982algorithms}. 
Given a trajectory $\Vec{p_t}$ with timestamps $t\in T$, we find a smooth spline approximation of degree 3 with smoothing condition $s$(=5.0) by ``scipy.interpolate.splrep''. The b-spline knot vector and control points of the approximated curve are $k$ and $\Vec{c}$. 
\begin{equation}
B(k, \Vec{c}) = F_{Bspline}([\Vec{p_t}], T, s)
\end{equation}

\mysubsubsection{Optimizing with floorplan} Given a b-spline knot vector $k$ and control points $\vec{c}$ of the approximated curve for synthetic data, and a walkable region, we optimize the location of control points, $\vec{c_m}$, using non-linear least squares to minimize the following energy function.
\begin{equation}
\mathop{\mathrm{\arg\min}}_{\vec{c_m}} \sum_{t \in T} f_{map}(B(k, \vec{c_m})(t)) + 2.0 \times \|B(k, \vec{c_m})(t) - \vec{p_t} \| \notag
    \label{eq:optimize}
\end{equation}
where $f_{map}$ is 
 a function of distance from a non-walkable pixel, which is computed by a flood-fill algorithm and smoothed by the Gaussian function ($\sigma=2.0$).
We sample the smoothed spline at constant distance of $1$ pixel with a Gaussian noise ($\sigma=3.0$) to obtain velocity vectors and ground-truth positions.

\subsection{Data Augmentation} 

We further perform the following three data augmentation tricks.
First, to ensure that motion directions are not restricted to a discrete set, we randomly rotate the walkable region image by ``scipy.ndimage.rotate'' function before generating synthetic trajectories by the shortest path algorithm.
Second, we add random perturbations to velocity magnitude and angular rate of the input 2D velocity vector sequence by the Gaussian function ($\sigma_{scale}=0.2$ pixels and $\sigma_{bias}=0.05$ radians per frame) to mimic scaling errors in inertial navigation and IMU gyroscope bias errors.
Third we perform a random in-plane rotation on the input velocity sequence to ensure the learned features are invariant to the unknown starting orientation alignment.

\begin{table*}[h]
\centering
\begin{tabular}{|l|l|l|l|l|}
\hline
Branch & Module & Layers/Operations & Input Shape & Output Shape \\ \hline
\multirow{11}{*}{\begin{tabular}[c]{@{}l@{}}Velocity \\ Branch\end{tabular}} & Velocity Input &  & {[}n, 2, 200{]} &  \\ \cline{2-5} 
 & \multirow{4}{*}{TCN Velocity Compressor} & TCN & {[}n, 2, 200{]} & {[}n, 288, 200{]} \\ \cline{3-5} 
 &  & Down-sample & {[}n, 288, 200{]} & {[}n, 288, 20{]} \\ \cline{3-5} 
 &  & Positional Encoding & {[}n, 288, 20{]} & {[}n, 432, 20{]} \\ \cline{3-5} 
 &  & Permute & {[}n, 432, 20{]} & {[}20, n, 432{]} \\ \cline{2-5} 
 & Shared Trans. Velocity Encoder&  Transformer Encoder & {[}20, n, 432{]} & {[}20, n, 432{]} \\ \cline{2-5} 
 & Transformer Velocity Encoder & Transformer Encoder & {[}20, n, 432{]} & {[}20, n, 432{]} \\ \cline{2-5} 
 & \multirow{4}{*}{TA-location decoder} & Permute & {[}20, n, 432{]} & {[}n$\times$20, 1, 24, 18{]} \\ \cline{3-5} 
 &  & Transpose-Conv & {[}n$\times$20, 1, 24, 18{]} & {[}n$\times$20, 3, 211, 157{]} \\ \cline{3-5} 
 &  & TA-Conv & {[}n$\times$20, 3, 211, 157{]} & {[}n$\times$20, 1, 211, 157{]} \\ \cline{3-5} 
 &  & Permute & {[}n$\times$20, 1, 211, 157{]} & {[}n, 20, 211, 157{]} \\ \hline\hline
\multirow{9}{*}{\begin{tabular}[c]{@{}l@{}}Location \\ Branch\end{tabular}} & Location input &  & {[}n, 20, 211,157{]} &  \\ \cline{2-5} 
 & \multirow{3}{*}{CNN Location Encoder} & Permute & {[}n, 20, 211, 157{]} & {[}n$\times$20, 1, 211, 157{]} \\ \cline{3-5} 
  &  & Conv & {[}n$\times$20, 1, 211, 157{]} & {[}n$\times$20, 1, 24, 18{]} \\ \cline{3-5} &  & Permute & {[}n$\times$20, 1, 24, 18{]} & {[}20, n, 432{]} \\ \cline{3-5} 
 &  & Positional Encoding & {[}20, n, 432{]} & {[}20, n, 432{]} \\ \cline{2-5} 
 & Transformer Location Decoder& Transformer Decoder  & {[}20, n, 432{]}, {[}20, n, 432{]} & {[}20, n, 432{]} \\ \cline{2-5} 
 & \multirow{4}{*}{TA-location decoder} & Permute & {[}20, n, 432{]} & {[}n$\times$20, 1, 24, 18{]} \\ \cline{3-5} 
 &  & Transpose-Conv & {[}n$\times$20, 1, 24, 18{]} & {[}n$\times$20, 3, 211, 157{]} \\ \cline{3-5} 
 &  & TA-Conv & {[}n$\times$20, 3, 211, 157{]} & {[}n$\times$20, 1, 211, 157{]} \\ \cline{3-5} 
 &  & Permute & {[}n$\times$20, 1, 211, 157{]} & {[}n, 20, 211, 157{]} \\ \hline
\end{tabular}
\caption{The input/output dimensions of our network for Building A. The tensors are of batch size $n$, and the location branch is shown with maximum sequence length 20 corresponding to velocity input sequence length of 200 frames. For buildings B and C, velocity input shape remains the same and location input/output shapes change to {[}n, 20, 144, 368{]} and {[}n, 20, 112, 384{]} resp., and the inner feature dimensions change accordingly.
%\yasu{Can you write how things are different for Building B and C? Maybe we wrote in the main paper but repeat here?}
}
\label{tbl:supple:network}
\end{table*}

\begin{table*}[tb]
\centering
\begin{subtable}{\linewidth}
\begin{tabular}{|l|l||rrrrrr||rrrrrr||c|}
\hline
\multirow{3}{*}{\begin{tabular}[c]{@{}l@{}}Buil-\\ ding\end{tabular}} & \multirow{3}{*}{Meth.} & \multicolumn{6}{c||}{Fixed short sequence (100 m)} & \multicolumn{6}{c||}{Full test sequence} & \multicolumn{1}{c|}{\multirow{3}{*}{\begin{tabular}[c]{@{}c@{}}run time\\cpu/gpu\\ (sec) $\downarrow$\end{tabular}}} \\ \cline{3-14}
 &  & \multicolumn{4}{c|}{SR(\%) at distance $\uparrow$} & \multicolumn{2}{c||}{SR(\%) at A  $\uparrow$} & \multicolumn{4}{c|}{SR(\%) at distance  $\uparrow$} & \multicolumn{2}{c||}{SR(\%) at A $\uparrow$} & \multicolumn{1}{c|}{} \\ \cline{3-14}
 &  & \multicolumn{1}{c}{1m} & \multicolumn{1}{c}{2m} & \multicolumn{1}{c}{4m} & \multicolumn{1}{c|}{6m} & \multicolumn{1}{c}{$20^{\circ}$} & \multicolumn{1}{c||}{$40^{\circ}$} & \multicolumn{1}{c}{1m} & \multicolumn{1}{c}{2m} & \multicolumn{1}{c}{4m} & \multicolumn{1}{c|}{6m} & \multicolumn{1}{c}{$20^{\circ}$} & \multicolumn{1}{c||}{$40^{\circ}$} & \multicolumn{1}{c|}{} \\ \hline\hline
 & PF & 22.3 & 43.2 & 59.4 & \multicolumn{1}{r|}{66.3} & 60.1 & 71.5 & 18.6 & 41.9 & 62.3 & \multicolumn{1}{r|}{68.2} & 62.1 & 71.8 &\second{1.4} /  5.8 \\
 & CRF &\second{26.2} &\second{52.9} & \best{76.2} & \multicolumn{1}{r|}{\best{87.8}} & \best{81.2} & \best{92.6} &\second{22.6} &\second{52.7} & \best{73.8} & \multicolumn{1}{r|}{\best{83.6}} & \best{79.7} & \best{90.8} & \ 9.2 /  \second{3.7} \\
\multirow{-3}{*}{A} & Ours & \best{34.4} & \best{59.9} &\second{74.5} & \multicolumn{1}{r|}{\second{ 81.2}} &\second{74.7} &\second{81.8} & \best{30.0} & \best{54.1} &\second{70.4} & \multicolumn{1}{r|}{\second{ 76.5}} &\second{70.2} &\second{77.9} & \best{ 0.3 /  0.1} \\ \hline
 & PF & 13.9 & 34.3 & 59.7 & \multicolumn{1}{r|}{71.2} & 59.1 & 73.4 & 15.4 & 39.7 & 58.2 & \multicolumn{1}{r|}{64.1} & 57.6 & 71.1 &\second{ \ 3.8 /  4.4} \\
 & CRF &\second{27.1} &\second{65.8} & \best{88.0} & \multicolumn{1}{r|}{\best{93.2}} & \best{88.1} & \best{93.9} &\second{23.7} &\second{62.8} & \best{87.1} & \multicolumn{1}{r|}{\best{91.2}} & \best{87.8} & \best{93.9} & 18.8 /  5.4 \\
\multirow{-3}{*}{B} & Ours & \best{47.6} & \best{69.3} &\second{74.5} & \multicolumn{1}{r|}{\second{ 77.3}} &\second{67.9} &\second{75.1} & \best{49.4} & \best{73.1} &\second{80.1} & \multicolumn{1}{r|}{\second{ 82.0}} &\second{72.7} &\second{80.7} & \best{1.2 / \ 0.2} \\ \hline
 & PF & 27.9 & 44.0 & 61.6 & \multicolumn{1}{r|}{72.1} & 27.6 & 47.4 & 25.3 & 37.1 & 50.6 & \multicolumn{1}{r|}{59.0} & 26.3 & 46.1 &\second{ 9.0 / 13.3} \\
 & CRF &\second{46.5} &\second{60.4} &\second{72.5} & \multicolumn{1}{r|}{\second{ 82.2}} &\second{45.9} &\second{64.9} &\second{48.7} &\second{65.4} &\second{77.2} & \multicolumn{1}{r|}{\second{ 85.2}} &\second{47.6} &\second{68.3} & 36.0 / 17.4 \\
\multirow{-3}{*}{C} & Ours & \best{70.7} & \best{78.7} & \best{84.1} & \multicolumn{1}{r|}{\best{87.6}} & \best{52.4} & \best{68.0} & \best{73.1} & \best{80.8} & \best{85.4} & \multicolumn{1}{r|}{\best{89.3}} & \best{53.6} & \best{69.9} & \best{2.4 /  \ \ 0.7} \\ \hline
\end{tabular}
\caption{Inertial Re-localization $\mathbb{R}^2$}
\label{tbl:results:relocalization:r2}
\end{subtable}\par

\bigskip

\begin{subtable}{\linewidth}
\begin{tabular}{|l|l||rrrrrr||rrrrrr||c|}
\hline
\multirow{3}{*}{\begin{tabular}[c]{@{}l@{}}Buil-\\ ding\end{tabular}} & \multirow{3}{*}{Meth.} & \multicolumn{6}{c||}{Fixed short sequence (100 m)} & \multicolumn{6}{c||}{Full test sequence} & \multicolumn{1}{c|}{\multirow{3}{*}{\begin{tabular}[c]{@{}c@{}}run time\\cpu/gpu\\ (sec) $\downarrow$\end{tabular}}} \\ \cline{3-14}
 &  & \multicolumn{4}{c|}{SR(\%) at distance $\uparrow$} & \multicolumn{2}{c||}{SR(\%) at A  $\uparrow$} & \multicolumn{4}{c|}{SR(\%) at distance  $\uparrow$} & \multicolumn{2}{c||}{SR(\%) at A $\uparrow$} & \multicolumn{1}{c|}{} \\ \cline{3-14}
 &  & \multicolumn{1}{c}{1m} & \multicolumn{1}{c}{2m} & \multicolumn{1}{c}{4m} & \multicolumn{1}{c|}{6m} & \multicolumn{1}{c}{$20^{\circ}$} & \multicolumn{1}{c||}{$40^{\circ}$} & \multicolumn{1}{c}{1m} & \multicolumn{1}{c}{2m} & \multicolumn{1}{c}{4m} & \multicolumn{1}{c|}{6m} & \multicolumn{1}{c}{$20^{\circ}$} & \multicolumn{1}{c||}{$40^{\circ}$} & \multicolumn{1}{c|}{} \\ \hline\hline
& PF & 23.6 & 42.3 & 60.2 & \multicolumn{1}{r|}{68.7} & 62.9 & 75.1 & 17.2 & 36.5 & 54.4 & \multicolumn{1}{r|}{61.0} & 56.9 & 67.6 & \second{ 0.6} /  3.1 \\
 & LP & 5.9 & 21.0 & 46.6 & \multicolumn{1}{r|}{59.8} & 55.0 & 78.5 & 4.8 & 19.8 & 40.1 & \multicolumn{1}{r|}{52.9} & 55.1 & 76.0 & 4.6 /  \second{0.3} \\
 & CRF & \second{ 27.5} & \second{ 55.0} & \best{ 77.7} & \multicolumn{1}{r|}{\best{ 88.8}} & \best{ 82.2} & \best{ 93.4} & \second{ 23.3} & \second{ 53.5} & \best{ 74.9} & \multicolumn{1}{r|}{\best{ 84.4}} & \best{ 80.9} & \best{ 92.5} & 9.5 /  3.7 \\
\multirow{-4}{*}{A} & Ours & \best{36.1} & \best{ 62.0} & \second{ 76.4} & \multicolumn{1}{r|}{\second{ 82.7}} & \second{ 76.7} & \second{ 83.1} & \best{ 31.2} & \best{ 56.0} & \second{ 73.0} & \multicolumn{1}{r|}{\second{ 79.8}} & \second{ 72.9} & \second{ 81.3} & \best{0.3 /   0.1} \\ \hline
 & PF & 14.9 & 36.1 & 62.8 & \multicolumn{1}{r|}{75.8} & 63.3 & 81.2 & 9.5 & 25.3 & 41.7 & \multicolumn{1}{r|}{48.4} & 46.8 & 61.5 & \second{ 1.4} /  4.4 \ \\
 & LP & 6.6 & 24.6 & 61.0 & \multicolumn{1}{r|}{73.8} & 62.2 & 80.4 & 2.2 & 9.9 & 26.9 & \multicolumn{1}{r|}{37.8} & 48.7 & 70.6 & 1.9 /  \second{0.8} \\
 & CRF & \second{ 29.7} & \best{ 71.2} & \best{ 92.3} & \multicolumn{1}{r|}{\best{ 96.3}} & \best{ 91.5} & \best{ 97.4} & \second{ 23.7} & \second{ 64.3} & \best{ 87.1} & \multicolumn{1}{r|}{\best{ 91.2}} & \best{ 87.8} & \best{ 93.9} & 18.8 /  5.3 \ \\
\multirow{-4}{*}{B} & Ours & \best{ 47.6} & \second{ 69.3} & \second{ 74.5} & \multicolumn{1}{r|}{\second{ 77.3}} & \second{ 67.9} & \second{ 75.1} & \best{ 49.4} & \best{ 73.1} & \second{ 80.1} & \multicolumn{1}{r|}{\second{ 82.0}} & \second{ 72.7} & \second{ 80.7} & \best{ 1.2 /  0.2} \\ \hline
 & PF & 30.1 & 46.1 & 65.4 & \multicolumn{1}{r|}{76.6} & 28.4 & 48.4 & 18.5 & 30.5 & 45.0 & \multicolumn{1}{r|}{55.4} & 21.8 & 38.9 & \second{ 4.3} / 12.7 \\
 & LP & 16.6 & 35.6 & 58.4 & \multicolumn{1}{r|}{76.9} & 30.4 & 50.4 & 6.8 & 15.8 & 29.3 & \multicolumn{1}{r|}{41.2} & 18.1 & 32.4 & 14.5 / \ \second{7.1} \\
 & CRF & \second{ 52.1} & \second{ 67.1} & \second{ 77.9} & \multicolumn{1}{r|}{\second{ 86.6}} & \second{ 49.0} & \second{ 69.0} & \second{ 48.7} & \second{ 65.4} & \second{ 77.2} & \multicolumn{1}{r|}{\second{ 85.1}} & \second{ 47.6} & \second{ 68.8} & 36.0 / 17.4 \\
\multirow{-4}{*}{C} & Ours & \best{ 71.4} & \best{ 79.3} & \best{84.4} & \multicolumn{1}{r|}{\best{87.7}} & \best{52.7} & \best{68.7} & \best{73.5} & \best{81.2} & \best{85.8} & \multicolumn{1}{r|}{\best{89.6}} & \best{53.9} & \best{ 70.2} & \best{ 2.3 / \ 0.7} \\ \hline
\end{tabular}
\caption{Inertial Re-localization $SE(2)$}
\label{tbl:results:relocalization:se2}
\end{subtable}
\caption{ 
Evaluation per building for re-localization task (Table 2.b of the main paper contains the average metrics across three buildings). 
We compare \shorttitles (ours) with three methods that require a floorplan as input: Particle Filter (PF), Learned Prior (LP) and Conditional Random Fields (CRF). We report success rate (SR) at a given error distance threshold and angle (A) threshold, per building. Run time is the average CPU or GPU time per 1 min of motion sequence. The best and second best results per column are shown in \best{orange} and \second{cyan}, respectively.}
\label{tbl:supple:reloc}
\end{table*}

\begin{table*}[h]
\centering
\begin{tabular}{l||rr||rrrrrr|}
\cline{2-9}
 & \multicolumn{2}{c||}{Velocity branch} & \multicolumn{6}{|c|}{Location branch} \\ \cline{2-9} 
 & \multicolumn{2}{|c||}{Localization} & \multicolumn{2}{c|}{Localization} & \multicolumn{2}{c|}{Reloc $\mathbb{R}^2$} & \multicolumn{2}{c|}{Reloc SE(2)} \\
\multicolumn{1}{r||}{SR(\%) at distance $\rightarrow$} & 2m & 4m & 2m & \multicolumn{1}{r|}{4m} & 2m & \multicolumn{1}{r|}{4m} & 2m & 4m \\ \hline
\multicolumn{1}{|l||}{w/o transformer encoder {[}LSTM{]}} & 30.5 & 43.8 & - & \multicolumn{1}{r|}{-} & - & \multicolumn{1}{r|}{-} & - & - \\
\multicolumn{1}{|l||}{w/o transformer encoder {[}TCN{]}} & 31.2 & 43.2 & - & \multicolumn{1}{r|}{-} & - & \multicolumn{1}{r|}{-} & - & - \\
% \multicolumn{1}{|l||}{w/o location branch} & 27.8 & 40.3 & - & \multicolumn{1}{r|}{-} & - & \multicolumn{1}{r|}{-} & - & - \\
\multicolumn{1}{|l||}{w/o scheduled sampling} & 25.3 & 38.1 & 9.3 & \multicolumn{1}{r|}{15.2} & 9.3 & \multicolumn{1}{r|}{15.2} & 9.4 & 15.6 \\
\multicolumn{1}{|l||}{w/o synthetic data} & 42.4 & 64.5 & 10.4 & \multicolumn{1}{r|}{18.9} & 14.1 & \multicolumn{1}{r|}{25.0} & 17.5 & 29.1 \\ \hline
\multicolumn{1}{|l||}{Ours} & 52.5 & 72.1 & 44.8 & \multicolumn{1}{r|}{62.6} & 54.1 & \multicolumn{1}{r|}{70.4} & 56.0 & 73.0 \\ \hline
\end{tabular}
\caption{Ablation Study: The first two rows are results after replacing transformer encoder module with different neural architectures, in particular LSTM and TCN. Third and fourth rows show the effectiveness of our training and data augmentation processes resp. 
The success rate (\%) at two distance thresholds (m) on building A are the metrics.}
\label{tbl:supple:ablation}
\end{table*}

%%%%%%%%% REFERENCES
{\small
\bibliographystyle{ieee_fullname}
\bibliography{egbib}
}